\documentclass[10pt,twocolumn,letterpaper]{article}

\usepackage{iccv}              

\usepackage[accsupp]{axessibility}  

\definecolor{iccvblue}{rgb}{0.21,0.49,0.74}
\usepackage[pagebackref,breaklinks,colorlinks,allcolors=iccvblue]{hyperref}

\usepackage{multirow}
\usepackage{bbding}
\usepackage{utfsym}
\usepackage{wasysym}
\usepackage{colortbl}

\usepackage{float}

\def\paperID{2902} 
\def\confName{ICCV}
\def\confYear{2025}

\title{DeepShield: Fortifying Deepfake Video Detection with  Local and Global Forgery Analysis}

\author{Yinqi Cai$^{1,2}$\thanks{Equal Contribution.}{\hspace{2.5em}}Jichang Li$^{2*}${\hspace{2.5em}}Zhaolun Li$^{3}${\hspace{2.5em}}Weikai Chen\thanks{This paper solely reflects the author's personal research and is not associated with the author's affiliated institution.}{\hspace{2.5em}}Rushi Lan$^{3}${\hspace{2.5em}}Xi Xie$^{1}$\\Xiaonan Luo$^{3}${\hspace{2.5em}}Guanbin Li$^{1,2,4}\thanks{Corresponding Author.}$ \vspace{0mm}\\
$^1$Sun Yat-sen University\quad $^2$Pengcheng Laboratory\quad $^3$Guilin University of Electronic Technology\quad \vspace{0mm}\\
$^4$Guangdong Key Laboratory of Big Data Analysis and Processing\\
{\tt\small caiyq27@mail2.sysu.edu.cn, li.jichang@pcl.ac.cn, liguanbin@mail.sysu.edu.cn}
 \vspace{0mm}
}

\begin{document}
\maketitle
\begin{abstract}
Recent advances in deep generative models have made it easier to manipulate face videos, raising significant concerns about their potential misuse for fraud and misinformation. Existing detectors often perform well in in-domain scenarios but fail to generalize across diverse manipulation techniques due to their reliance on forgery-specific artifacts. In this work, we introduce DeepShield, a novel deepfake detection framework that balances local sensitivity and global generalization to improve robustness across unseen forgeries. DeepShield enhances the CLIP-ViT encoder through two key components: Local Patch Guidance (LPG) and Global Forgery Diversification (GFD). LPG applies spatiotemporal artifact modeling and patch-wise supervision to capture fine-grained inconsistencies often overlooked by global models. GFD introduces domain feature augmentation, leveraging domain-bridging and boundary-expanding feature generation to synthesize diverse forgeries, mitigating overfitting and enhancing cross-domain adaptability. Through the integration of novel local and global analysis for deepfake detection, DeepShield outperforms state-of-the-art methods in cross-dataset and cross-manipulation evaluations, achieving superior robustness against unseen deepfake attacks. Code is available at https://github.com/lijichang/DeepShield.
\end{abstract}
    
\vspace{-5pt}
\section{Introduction}

Rapid advancement of deep generative models, including GANs~\cite{goodfellow2020gan,karras2019stylegan}, VAEs~\cite{kingma2013vae,lai2025llm}, and diffusion models~\cite{dhariwal2021diffusion,zhang2024UniFL}, has significantly lowered the barrier for manipulating facial videos, raising serious concerns about their misuse in fraud, misinformation, and privacy violations. 
As a result, robust and generalizable deepfake detection methods have become a critical research focus to safeguard digital media integrity.

\begin{figure}[!t]
\centering
\includegraphics[width=1.0\columnwidth]{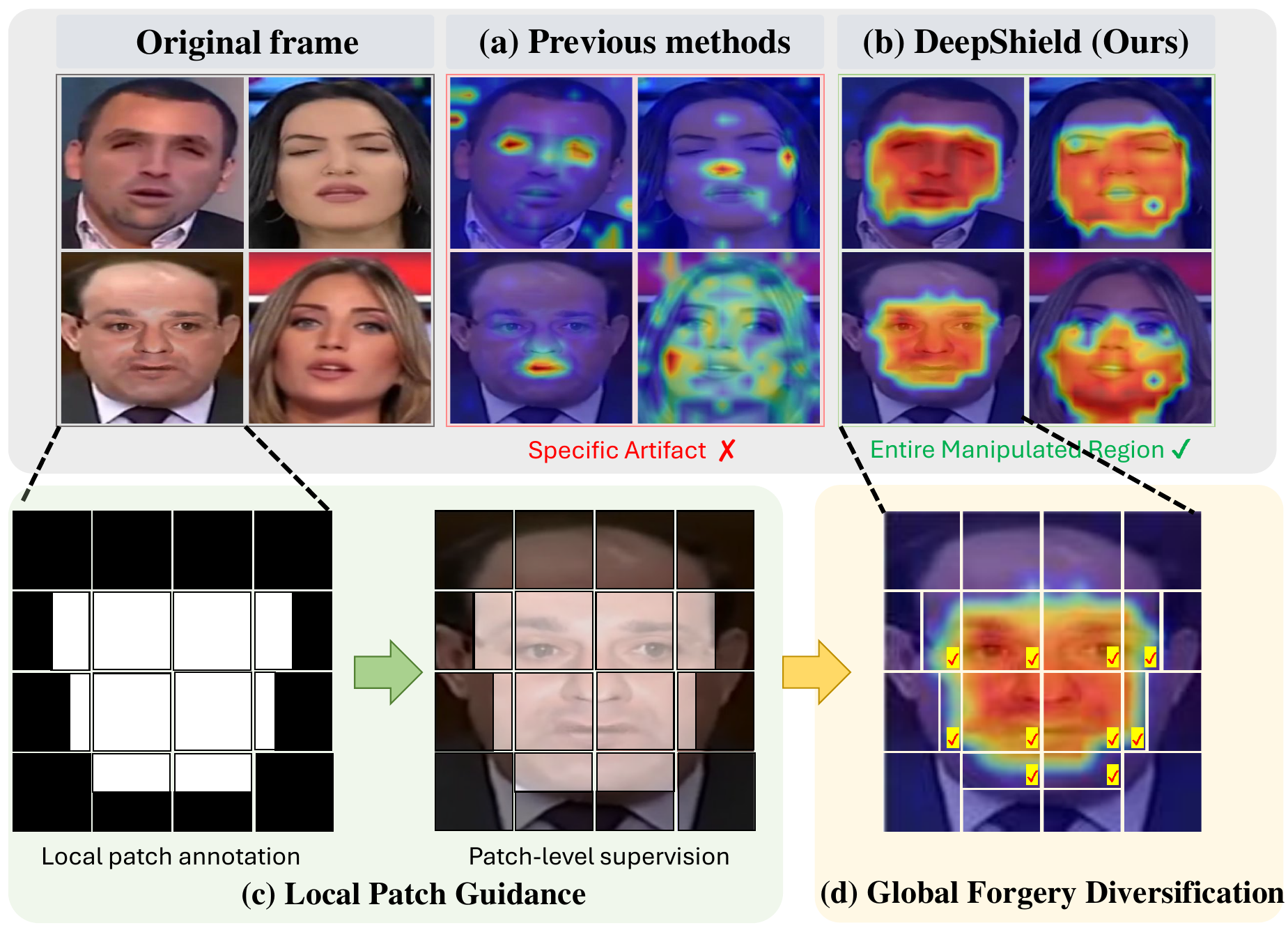} 
\caption{
(a)-(b) Comparison of our DeepShield and previous methods in artifact localization using patch-based attention heatmaps. These heatmaps show attention responses from CLIP-ViT for randomly selected image patches. While previous methods emphasize only the most salient specific artifacts using global features, our DeepShield allows patch tokens to capture nuanced details across entire manipulated facial regions.
(c)-(d) Illustration of DeepShield’s local-to-global learning paradigm: Local Patch Guidance, and Global Forgery Diversification.
} 
\label{fig:intro}
\end{figure}

While prior deepfake detectors~\cite{liu2020global,zhao2021multi,jeong2022frepgan,shuai2023locate} show strong performance in in-domain evaluations, they often fall short in cross-domain scenarios where manipulation techniques differ from those seen at training.
Existing approaches~\cite{li2020xray,chen2022SLADD,shiohara2022sbi,larue2023seeable} attempt to address the generalization challenge by retraining the model on an augmented dataset with deepfake videos produced with a variety of manipulation techniques. However, this strategy is inherently time-consuming, computationally expensive, and limited in scalability, as it requires continuous updates to keep pace with emerging deepfake generation methods.
More recently, large-scale vision-language models like CLIP~\cite{CLIP} have been adopted for deepfake detection due to their strong generalization capabilities~\cite{han2024towards,yan2025generalizing}. However, these models are primarily optimized for image-level tasks and tend to rely heavily on global features, making them less effective at capturing fine-grained forgery traces.
Moreover, these models tend to concentrate on prominent artifacts within manipulated regions, while overlooking subtle forgery signals that are crucial for identifying more sophisticated deepfakes. 
As illustrated in Figure \ref{fig:intro} (a), these models exhibit strong focus on salient, globally pronounced artifacts in manipulated videos, such as exaggerated transitions in facial regions, while failing to capture subtle forgery traces such as blending boundaries.
This tendency to prioritize salient features over nuanced inconsistencies can cause models to overfit to specific forgery types seen during training, further reducing their effectiveness when exposed to novel, cross-domain manipulations. 

To address these challenges, we introduce \textit{DeepShield}, a novel framework designed to enhance the model generalization while maintaining sensitivity to both subtle and pronounced manipulations. 
Towards this goal, DeepShield employs a local-to-global learning paradigm that combines granular patch analysis with diverse forgery representation. 
At the core of our method are two complementary components: Local Patch Guidance (LPG) and Global Forgery Diversification (GFD), with its conceptual pipeline as illustrated in Figure \ref{fig:intro} (c) and (d). Our core idea is to implement deep coupling and iterative collaboration between the proposed LPG and GFD, thus creating a dynamic framework  capable of precisely emphasizing local forgery details and effectively uncovering unknown global forgery patterns.

In particular, LPG enhances the model's sensitivity to local forgery inconsistencies by enforcing patch-level supervision, enabling it to capture subtle manipulation patterns that global feature-based approaches often overlook.
By establishing stronger semantic correlations between patches, LPG refines the model's ability to detect fine-grained anomalies, significantly improving detection accuracy.
As illustrated in Figure \ref{fig:intro} (b), {LPG} successfully identifies minute inconsistencies, such as blending boundaries, across local patches. This balanced attention to both local and global features improves prediction accuracy and finally leads to robust detection of the entire manipulated regions in the deepfake videos. 
Additionally, we introduce Spatiotemporal Artifact Modeling (SAM) to blend novel deepfake videos with spatial artifacts and temporal inconsistencies. This addresses the lack of local annotations in original training videos while enriching the model with rich and diverse training samples, further improving its generalization capability.

On the other hand, {GFD} addresses forgery-specific overfitting and enhances the model’s generalization across diverse deepfake domains. 
Specifically, GFD synthesizes a broad spectrum of forgery data, enhancing the model’s capabilities through targeted training on this enriched dataset.
This data synthesis is achieved by means of domain bridging and boundary expanding.
As illustrated in Figure~\ref{fig:GFD} (a), Domain-Bridging Feature Generation (DFG) bridges domain gaps by blending distributions from different forgery types, helping the model learn domain-invariant features and improving its adaptability across various manipulation styles.
In the meanwhile, Boundary-Expanding Feature Generation (BFG) further broadens the model's scope by generating features at the boundary of existing forgery domains 
(See~Figure~\ref{fig:GFD} (b)).
By scaling the standard deviation of each feature channel, BFG extends the model's detection range to include more subtle and complex forgery types without overlapping with real data clusters.
We additionally design a dedicated training objective applied to original and generated samples, combining standard cross-entropy loss with supervised contrastive loss~\cite{khosla2020supcon}. 
This objective enables the model to distinguish between real and fake features while reinforcing domain invariance among diverse forgery techniques.

Experimental results demonstrate that LPG and GFD work in tandem to equip DeepShield with the ability to address both local and global challenges in deepfake video detection, resulting in substantial performance improvements, particularly in cross-domain detection scenarios.
We summarize our contributions as follows:
\begin{itemize}
    \item We propose DeepShield, a novel framework for deepfake video detection that combines local and global analysis to improve generalization across diverse manipulation techniques.
    \item We introduce {Local Patch Guidance (LPG)}, which enhances the model’s ability to capture local and fine-grained forgery inconsistencies by explicitly constructing spatiotemporal patches masks to provide structured supervision signals.
    \item We present {Global Forgery Diversification (GFD)}, which addresses forgery-specific overfitting and enhances cross-domain generalization by synthesizing diverse forgery representations and applying a novel training strategy with supervised contrastive loss.
    \item Extensive experiments have demonstrated that the combination of LPG and GFD significantly improves the performance of detecting deepfake videos over existing state-of-the-art algorithms, especially in cross-domain settings.
\end{itemize}

\begin{figure*}[t]
\centering
\includegraphics[width=0.9\textwidth]{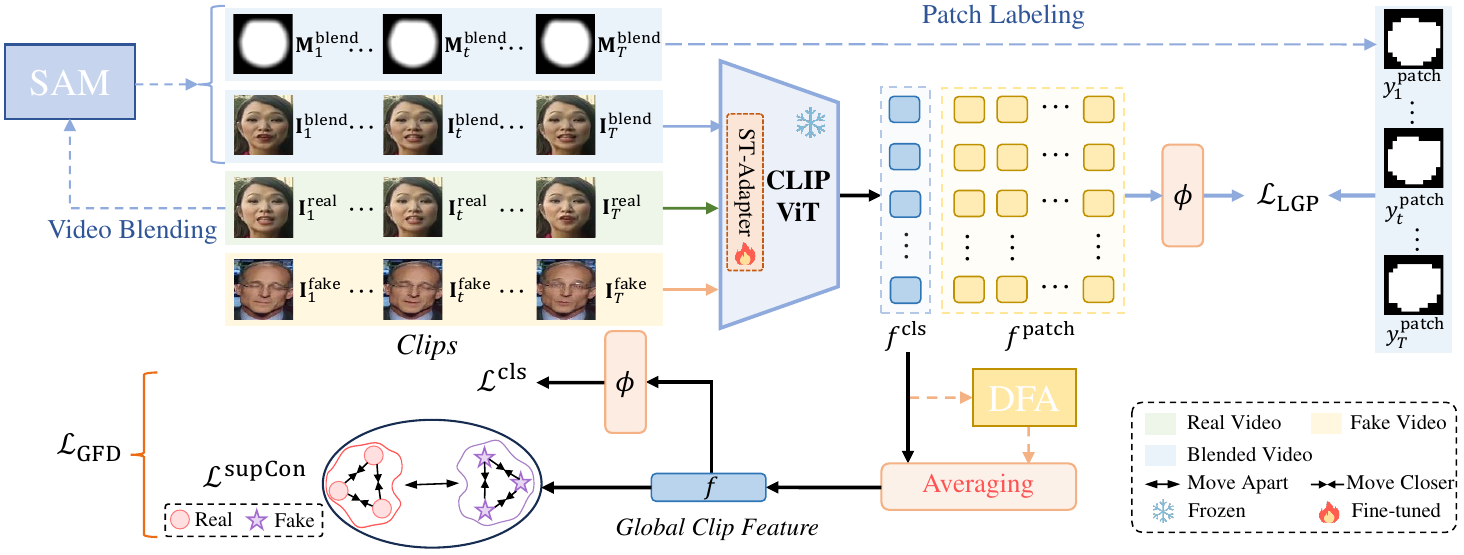}
\caption{
An overview of the DeepShield framework for deepfake video detection. The framework integrates Local Patch Guidance (LPG) and Global Forgery Diversification (GFD) to enhance generalization across diverse manipulation techniques. LPG improves sensitivity to subtle forgery inconsistencies by applying supervised learning on individual video patches, with the proposed Spatiotemporal Artifact Modeling (SAM) to blend deepfake videos. GFD addresses forgery-specific overfitting and strengthens cross-domain generalization by synthesizing diverse forgery representations through Domain Feature Augmentation (DFA). A dedicated training objective, combining standard cross-entropy loss with supervised contrastive loss, is employed to optimize detection performance. This unified approach enables the model to effectively capture both local and global manipulation traces, ensuring robust detection across various deepfake domains.
} 
\label{fig:framework}
\end{figure*}
\section{Related work}
\label{sec:related}

\paragraph{Deepfake Detection}
Deepfake detection methods are typically categorized into two main types based on the input modality: image-level and video-level detectors. Image-level methods focus on identifying spatial forgery traces, such as subtle facial artifacts~\cite{rossler2019ff++,liu2020global,zhao2021multi,ba2024exposing}, inconsistencies in identity information~\cite{dong2022ict,huang2023implicit}, and frequency-based anomalies~\cite{jeong2022frepgan,shuai2023locate}.
In contrast, video-level detectors leverage temporal information, analyzing multiple frames to capture dynamic inconsistencies essential for deepfake detection. For instance, STIL~\cite{gu2021stil} integrates two modules to detect both spatial and temporal inconsistencies, while LipForensics~\cite{haliassos2021lips} focuses on temporal inconsistencies in mouth movements. AltFreezing~\cite{wang2023altfreezing} alternates training spatial and temporal convolutional layers to comprehensively capture spatiotemporal information.
These approaches highlight the importance of combining spatial and temporal cues but typically rely on training from scratch with limited labeled data, restricting their generalization to novel forgery types. 
Unlike existing methods, our proposed DeepShield leverages pre-trained vision representations and synthesizes generalized spatiotemporal artifacts to generate richer supervision signals for local patch learning, significantly boosting detection performance in cross-domain.
\paragraph{Deepfake Detection Using Synthetic Data}
Drawing inspiration from the blending operations used in deepfake generation, several methods aim to synthesize new training frames with generalized spatial artifacts, such as blending traces and statistical inconsistencies. For example, Face X-Ray~\cite{li2020xray} blends two frames with similar facial landmarks to reproduce blending artifacts. SBI~\cite{shiohara2022sbi} enhances this by blending a frame with its augmented variant, introducing more realistic blending to improve the model's generalization and make detection more challenging. SeeABLE~\cite{larue2023seeable} employs a more subtle approach by blending only small portions of the face, helping the model detect finer inconsistencies. However, these methods tend to overlook temporal discontinuities that arise between frames during deepfake generation, as well as deepfakes that lack prominent spatial artifacts.  To address these limitations, AltFreezing~\cite{wang2023altfreezing} generates fake clips with temporal inconsistencies by randomly dropping or repeating frames, but this does not accurately mimic the actual deepfake creation process. 
Moreover, some studies~\textcolor{black}{~\cite{yan2025generalizing, li2023stcatcher, lin2024fake} attempt to blend video frames using landmark misalignment or adjacent-frame inconsistencies, but these methods often introduce unrealistic artifacts significantly different from real scenarios.} 

Different from these previous algorithms, we propose to synthesize videos with more natural spatiotemporal manipulation traces through mimicking the generation patterns of deepfake videos, aiming to achieve better detection of generalized artifacts. Additionally, recent works~\cite{cheng2024can, cheng2025stacking} attempt to diversify forgery samples in feature space but utilize simple linear interpolation between features from different domains, restricting the richness of feature representation. In contrast, this study introduces a distribution-based feature mixing and expanding, enabling more complex nonlinear augmentation to further enhance the diversity of forgery samples in feature space and improve the generalization capability of detection models.
\section{Methodology}
\label{sec:method}

\subsection{Preliminaries}

\paragraph{Overview}
In this work, we propose \textit{DeepShield}, a novel framework for deepfake video detection that combines local and global analysis to improve generalization. DeepShield consists of two key components: Local Patch Guidance (LPG) and Global Forgery Diversification (GFD). Specifically, LPG applies supervised learning to individual video patches, improving the model's sensitivity to local subtle forgery traces. In the meanwhile, GFD synthesizes diverse forgery representations and performs global feature learning, addressing forgery-specific overfitting and thus leading to enhanced cross-domain generalization. 
An overview of DeepShield can be found in~Figure~\ref{fig:framework}.

\vspace{-10pt}
\paragraph{Input and Output Configuration}
Let $\mathbf{V} \in \mathbb{R}^{T \times H \times W \times 3}$ represent a video clip consisting of $T$ consecutive cropped face frames, where $H$ and $W$ denote the height and width of each frame, respectively. We feed $\mathbf{V}$ into the video encoder $E$, obtaining the class embedding $f_t^{\text{cls}} \in \mathbb{R}^{C}$ for each frame $t = 1, \cdots, T$ and their corresponding patch embeddings $f_{t,p}^{\text{patch}} \in \mathbb{R}^{C}$ for $p = 1, \cdots, P$, where $P$ is the number of patch tokens per frame, and $C$ is the embedding dimension. In this study, we utilize CLIP's pre-trained image encoder, referred to as CLIP-ViT, as the video encoder $E$. To capture spatial manipulations and temporal inconsistencies efficiently, we fine-tune CLIP-ViT with the parameter-efficient ST-Adapter~\cite{pan2022stada}. The ST-Adapter is inserted before the Multi-Head Self-Attention and Feed-Forward Network in each Transformer block of CLIP-ViT.

\subsection{Local Patch Guidance}
\label{LPG}

To improve the model's sensitivity to local forgery anomalies, we propose {Local Patch Guidance (LPG)}, which leverages patch-level supervised learning to strengthen the semantic correlations among patches. A key challenge is that both real and deepfake videos typically lack precise annotations of manipulated areas, making direct patch-level supervision impractical. To address this, we introduce \textit{Spatiotemporal Artifact Modeling (SAM)}, which generates deepfake videos with manipulated masks by blending real videos. This scheme creates precisely labeled training data that highlights generalized spatiotemporal artifacts, facilitating patch-level learning to scale to novel forgery data.

\noindent
\paragraph{Generating Patch-Level Annotations} 
After obtaining blended fake videos via SAM and retaining original real videos, we divide each video frame and its corresponding manipulation mask into $P$ non-overlapping flattened patches. Let a single frame be of size $H \times W$. Each patch is then of size $H_p = \frac{H}{\sqrt{P}}$ and $W_p = \frac{W}{\sqrt{P}}$. This {Patch Division} step transforms every frame (and mask) into a collection of smaller local regions, allowing us to assign patch-level forgery labels.

To determine whether a given patch $p$ is “real” or “fake”, we define a {\textit{Patch Scoring Function}} that calculates the number of manipulated pixels within the corresponding mask:
{\setlength{\abovedisplayskip}{5pt}
\setlength{\belowdisplayskip}{5pt}
\begin{equation}
\text{PatchMaskScore}(\mathbf{M}_{t,p}) 
= \sum_{h=1}^{H_p} \sum_{w=1}^{W_p} \mathbb{I}\big[\mathbf{M}_{t,p}^{(h,w)} > 0\big],
\end{equation}}
where $\mathbf{M}_{t,p}$ is the manipulation mask of patch $p$ in frame $t$, $\mathbf{M}_{t,p}^{(h,w)}$ is the pixel value at the $h$-th row and $w$-th column of $\mathbf{M}_{t,p}$, and $\mathbb{I}[\cdot]$ is an indicator function returning 1 if the value is greater than 0 (manipulated region), and 0 otherwise. Based on a threshold $\theta$, we assign the patch label as:
{\setlength{\abovedisplayskip}{5pt}
\setlength{\belowdisplayskip}{5pt}
\begin{equation}
\label{eq:patch}
\text{y}_{t,p}^{\text{patch}} =
\begin{cases} 
    1, & \text{if } \text{PatchMaskScore}(\mathbf{M}_{t,p}) \geq \theta, \\
    0, & \text{otherwise}.
\end{cases}
\end{equation}}
Thus, patches in real videos are labeled 0, while patches in blended deepfake videos are labeled 1 only if they contain enough manipulated pixels. This scheme provides patch-level labels for training without needing manual local annotations.

\noindent
\paragraph{Learning Forgery Features at the Patch Level} 
Once patches are labeled, each patch token serves as an independent training sample in a binary classification task. Concretely, we feed the patch feature $f_{t,p}^{\text{patch}}$ (extracted by the video encoder) into a binary classifier $\phi$ to estimate $\text{Prob}_{t,p}^{\text{patch}}$, the probability of that patch being fake. The training objective is a binary cross-entropy loss over all patches:
{\setlength{\abovedisplayskip}{5pt}
\setlength{\belowdisplayskip}{5pt}
\begin{equation}
\begin{aligned}
    \mathcal{L}_{\text{LGP}} = -\frac{1}{T \cdot P} &\,\sum_{t,p} \Bigl[
        \text{y}_{t,p}^{\text{patch}} \cdot \log\!\bigl(\text{Prob}_{t,p}^{\text{patch}}\bigr)\\[-3pt]
        &\,+ \bigl(1 - \text{y}_{t,p}^{\text{patch}}\bigr) \cdot \log\!\bigl(1 - \text{Prob}_{t,p}^{\text{patch}}\bigr)
    \Bigr].
\end{aligned}
\end{equation}}
By separately supervising each patch token, the model is guided to cluster patch embeddings of the same category (real or fake) more closely and to separate different ones. This {enhances semantic correlations between patches}, because the model must learn how local features interact and differ within the broader video context.

\begin{figure}[t]
\centering
\includegraphics[width=1.0\columnwidth]{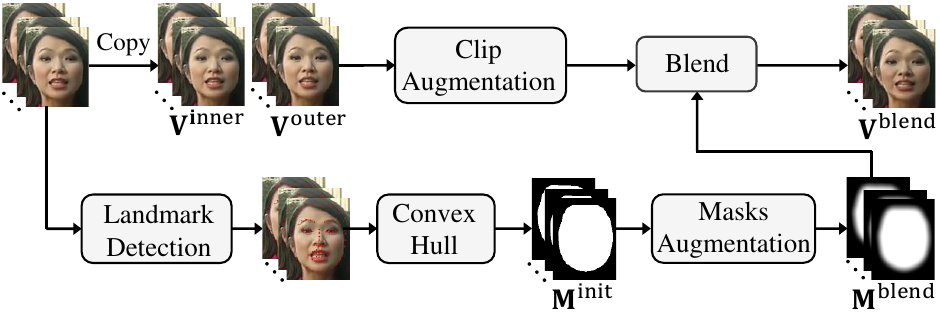} 
\caption{
An illustration of the Spatiotemporal Artifact Modeling (SAM), which synthesizes deepfake video clips with generalized spatiotemporal artifacts to facilitate local feature learning.
} 
\label{fig:sam}
\end{figure}
In the model’s self-attention mechanism, the {class token} attends to every patch token~\cite{dosovitskiy2020vit}. When patch embeddings become more discriminative, the class token can aggregate richer local semantics to detect subtle forgeries. As a result, {local forgery traces} are more effectively perceived, boosting the overall detection performance.

\noindent
\paragraph{Spatiotemporal Artifact Modeling (SAM)}
\label{SAM}
Deepfake videos often exhibit both spatial and temporal inconsistencies. To emulate these characteristics and create manipulated masks for patch-level supervision, we propose SAM, illustrated in Figure~\ref{fig:sam}. SAM extends the SBI technique~\cite{shiohara2022sbi} (originally designed for image-level blending) to video-level blending, introducing spatial artifacts and temporal inconsistencies in two ways:

\noindent
\emph{(1) Spatial Artifact Generating.} 
For each frame $\mathbf{I}_{t}$ in a video clip $\{\mathbf{I}_{t} \mid t=1,\cdots,T\}$, we produce an inner frame $\mathbf{I}^{\text{inner}}_{t}$ (providing the face) and an outer frame $\mathbf{I}^{\text{outer}}_{t}$ (providing the background) from the same original video frame by applying different enhancements (e.g., color, brightness, or sharpness) to one of them, thereby creating statistical discrepancies. We then blend the face region from $\mathbf{I}^{\text{inner}}_{t}$ into $\mathbf{I}^{\text{outer}}_{t}$ via a mask $\mathbf{M}_{t}^{\text{blend}}$ derived from the convex hull of predicted facial landmarks~\cite{king2009dlib} with random deformations and blurring. This yields a {blended} frame:
{\setlength{\abovedisplayskip}{5pt}
\setlength{\belowdisplayskip}{5pt}
\begin{equation}
\label{eq:sbi}
\mathbf{I}^{\text{blend}}_{t} 
= \mathbf{M}_{t}^{\text{blend}} \odot \mathbf{I}^{\text{inner}}_{t}
+ (1-\mathbf{M}_{t}^{\text{blend}}) \odot \mathbf{I}^{\text{outer}}_{t},
\end{equation}}
where $\odot$ denotes the Hadamard product. By blending statistically inconsistent face and background regions, spatial artifacts are introduced.

\noindent
\emph{(2) Temporal Artifact Generating.}
To model temporal inconsistencies across frames, SAM applies this blending procedure across all $T$ frames by mimicking the generation patterns of deepfake videos. Key considerations include: 1) one version (either $\mathbf{I}^{\text{inner}}_{t}$ or $\mathbf{I}^{\text{outer}}_{t}$) undergoes consistent enhancement across $T$ frames to prevent large disparities that could cause the model to overfit to unnatural forgeries; and 2) the random deformation and blurring of $\mathbf{M}_{t}^{\text{blend}}$ are consistently applied across $T$ frames, preserving slight differences across frames.

In sum, SAM produces deepfake video clips that contain more generalized spatiotemporal artifacts, offering high-quality manipulated masks for patch-level supervision. By training with these blended videos under our {Local Patch Guidance}, the model gains a finer-scale perception of forgery traces, significantly improving its capacity to detect local manipulations that may otherwise be overlooked.

\begin{figure}[!t]
\centering
\includegraphics[width=0.9\columnwidth]{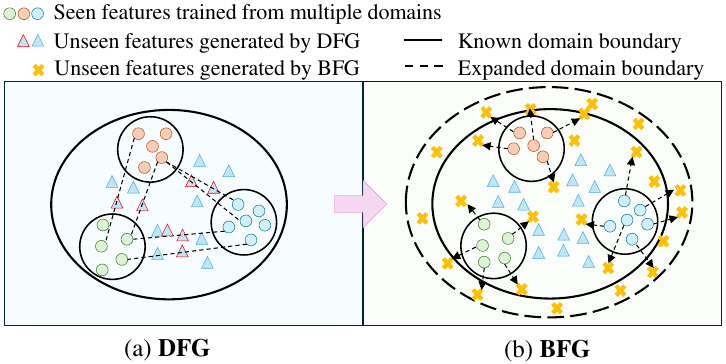} 
\caption{
An illustration to exemplify our proposed DFA for Domain-Bridging Feature Generation (DFG) and Boundary-Expanding Feature Generation (BFG).
} 
\label{fig:GFD} 
\end{figure}

\subsection{Global Forgery Diversification}
\label{GFD}

To enhance the model's generalization capability, we propose {Global Forgery Diversification (GFD)}. This module synthesizes diverse and domain-invariant forgery representations through \textit{Domain Feature Augmentation (DFA)} and employs a novel training strategy that combines cross-entropy loss with supervised contrastive loss. By mitigating forgery-specific overfitting, GFD ultimately improves the model's ability to detect previously unseen forgery types.

\vspace{-10pt}
\paragraph{Domain Feature Augmentation (DFA)}
\label{DFA}
To diversify forgery features, we introduce {Domain Feature Augmentation}, which generates novel forgery embeddings using two synthesis strategies: {Domain-Bridging Feature Generation (DFG)} and {Boundary-Expanding Feature Generation (BFG)}. In contrast to the previous approach~\cite{Yan2024LSDA} that typically relies on linear feature augmentation, DFA exploits distributional characteristics to expose the model to a wider range of potential forgery variations, thereby enhancing its cross-domain generalization. An illustration of the proposed DFA is provided in Figure~\ref{fig:GFD}.

\noindent
(1) \emph{Domain-Bridging Feature Generation (DFG).} 
Different forgery types exhibit domain gaps that hinder cross-domain generalization. Inspired by~\cite{zhou2021mixstyle}, {DFG} mixes distributions from different forgery domains to help the model learn domain-invariant representations. Concretely, we randomly select $N$ clips (each containing $T$ consecutive frames) from a deepfake video in a mini-batch. For each video, we compute the mean $\mu_{c}$ and standard deviation $\sigma_{c}$ of the class embeddings for each channel $c$:
{\setlength{\abovedisplayskip}{5pt}
\setlength{\belowdisplayskip}{5pt}
\begin{gather}
    \mu_{c} = \frac{1}{N \cdot T} \sum_{n=1}^{N} \sum_{t=1}^{T} f^{\text{cls}}_{n,t,c}, \\
    \sigma_{c} = \sqrt{\frac{1}{N \cdot T} \sum_{n=1}^{N} \sum_{t=1}^{T} \bigl( f^{\text{cls}}_{n,t,c} - \mu_{c} \bigr)^2}.
\end{gather}}
Next, we randomly pair each deepfake video with another from a different forgery type within the mini-batch. Let $\tilde{\mu}_{c}$ and $\tilde{\sigma}_{c}$ denote the paired video's mean and standard deviation, respectively. We sample a mixing weight $\lambda \sim Beta(0.1, 0.1)$ and interpolate:
{\setlength{\abovedisplayskip}{5pt}
\setlength{\belowdisplayskip}{5pt}
\begin{equation}
\begin{aligned}
\mu_{c}^{\text{mix}} &= \lambda \mu_{c} + (1 - \lambda) \tilde{\mu}_{c}, \\
\sigma_{c}^{\text{mix}} &= \lambda \sigma_{c} + (1 - \lambda) \tilde{\sigma}_{c}.
\end{aligned}
\end{equation}}
Finally, we apply AdaIN~\cite{huang2017adain} to generate domain-bridging features:
{\setlength{\abovedisplayskip}{5pt}
\setlength{\belowdisplayskip}{5pt}
\begin{equation}
    \hat{f}^{\text{cls}}_{n,t,c} 
    = \sigma_{c}^{\text{mix}} \cdot \biggl( \frac{f^{\text{cls}}_{n,t,c} - \mu_{c}}{\sigma_{c}} \biggr) + \mu_{c}^{\text{mix}}.
\end{equation}}
This procedure “bridges” domain gaps among different forgery types, enabling more robust, domain-invariant representations.

\noindent
(2) \emph{Boundary-Expanding Feature Generation (BFG).}
While domain-bridging features address known forgery spaces, they do not necessarily cover all potential variations. {BFG} tackles this by synthesizing features that lie on the outer boundaries of existing domain clusters, thereby further enlarging the forgery domain space. Concretely, we scale the standard deviation of each channel $c$ to slightly shift the features outward:
{\setlength{\abovedisplayskip}{5pt}
\setlength{\belowdisplayskip}{5pt}
\begin{equation}
    \bar{\sigma}_{c} = \alpha \cdot \sigma_{c},
\end{equation}}
where $\alpha > 1$ (set to $\alpha = 1.1$ in practice) offsets the features beyond the known forgery regions. The {boundary-expanding features} are then generated via:
{\setlength{\abovedisplayskip}{5pt}
\setlength{\belowdisplayskip}{5pt}
\begin{equation}
    \bar{f}^{\text{cls}}_{n,t,c} 
    = \bar{\sigma}_{c} \cdot \biggl( \frac{f^{\text{cls}}_{n,t,c} - \mu_{c}}{\sigma_{c}} \biggr) 
    + \mu_{c},
\end{equation}}
which expands the forgery boundaries without overlapping real feature clusters. By synthesizing such out-of-distribution examples, BFG promotes a decision space that is more generalized against diverse forgeries.

\noindent
\paragraph{Global Clip Feature Representation}
For each video clip $v$, we compute its global clip feature $f_v$ by averaging the frame-level class embeddings:
{\setlength{\abovedisplayskip}{5pt}
\setlength{\belowdisplayskip}{5pt}
\begin{equation}
    f_v = \frac{1}{T} \sum_{t=1}^{T} f^{\text{cls}}_{v,t}.
\end{equation}}
This aggregated representation captures both spatial and temporal cues essential for generalized forgery detection.

\paragraph{Dedicated Training Objective}
The objective in GFD promotes discrimination between real and fake samples while encouraging domain invariance across diverse forgery types. Formally, the total loss $\mathcal{L}_{\text{GFD}}$ is composed of a cross-entropy loss $\mathcal{L}^{\text{cls}}$ and a supervised contrastive loss $\mathcal{L}^{\text{supCon}}$:
{\setlength{\abovedisplayskip}{5pt}
\setlength{\belowdisplayskip}{5pt}
\begin{equation}
    \mathcal{L}_{\text{GFD}} = \mathcal{L}^{\text{cls}} + \upsilon\mathcal{L}^{\text{supCon}},
\end{equation}}
where $\upsilon$ is a weight balancing the two terms. During training, the model observes original training videos, blended videos from SAM (Section~\ref{SAM}), and features synthesized by DFA. This broad range of forgery instances compels the model to learn discriminative yet general features. Minimizing $\mathcal{L}_{\text{GFD}}$ sharpens the global representations of both real and fake videos, enhancing cross-domain deepfake detection.
To be more specific, the definitions of the loss functions $\mathcal{L}^{\text{cls}}$ and $\mathcal{L}^{\text{supCon}}$ can be summarized as follows.

    \noindent (1) \textit{Cross-Entropy Loss (}$\mathcal{L}^{\text{cls}}$\textit{)}. 
    This loss separates real from fake samples: 
    {\setlength{\abovedisplayskip}{5pt}
    \setlength{\belowdisplayskip}{5pt}
    \begin{equation}
    \mathcal{L}^{\text{cls}} = \frac{1}{B} \sum_{v=1}^{B} \text{H}(\text{Prob}_v, \text{y}_v),
    \end{equation}}
    where $\text{H}$ denotes the standard cross-entropy function, $B$ is the total number of original and generated samples in a batch, $\text{Prob}_v$ is the predicted probability for sample $v$, and $\text{y}_v$ is the ground-truth label.

    \noindent (2) \textit{Supervised Contrastive Loss (}$\mathcal{L}^{\text{supCon}}$\textit{)}. Following~\cite{khosla2020supcon}, this loss encourages intra-class compactness and inter-class dispersion:
    {\setlength{\abovedisplayskip}{5pt}
    \setlength{\belowdisplayskip}{5pt}
    \begin{equation}
        \mathcal{L}^{\text{supCon}} = \frac{1}{B} \sum_{v=1}^{B} -\frac{1}{|\text{J}(v)|} \sum_{i \in \text{J}(v)} L(v, i),
    \end{equation}}
    {\setlength{\abovedisplayskip}{5pt}
    \setlength{\belowdisplayskip}{5pt}
    \begin{equation*}
        L(v, i) = \log \frac{\exp(f_v \cdot f_i / \tau)}{\sum_{j \in \text{J}(v) \setminus \{ v \}} \exp(f_v \cdot f_j / \tau)},
    \end{equation*}}
    where $\text{J}(v)$ is the set of samples sharing the same class as sample $v$, $f_i$ and $f_j$ are the global clip features for samples $i$ and $j$, and $\tau$ is a temperature parameter. 

\paragraph{Model Optimization}
We combine $\mathcal{L}_{\text{LPG}}$ and $\mathcal{L}_{\text{GFD}}$ to fine-tune the CLIP-ViT model:
{\setlength{\abovedisplayskip}{5pt}
\setlength{\belowdisplayskip}{5pt}
\begin{equation}
    \mathcal{L}^{\text{overall}} =\omega\mathcal{L}_{\text{LPG}} + \mathcal{L}_{\text{GFD}},
    \label{eq:totalloss}
\end{equation}}
where $\omega$ balances the trade-off between local patch-level supervision and global forgery diversification. By training with this combined objective, the model gains robust detection performance across both local anomalies and diverse global forgeries.
\section{Experiments}
\label{sec:exp}

\begin{table*}[t]
\centering
\scriptsize
\begin{tabular}{lccccccc}
\toprule
\multirow{2}{*}{Method} & \multirow{2}{*}{Architecture} & \multirow{2}{*}{Input Type} & \multicolumn{5}{c}{Testing Set AUC (\%)} \\
\cmidrule{4-8}
& & & FF++ & CDF & DFDCP & DFDC & DFD \\
\midrule
Xception~\cite{rossler2019ff++} & Xception & Frame & 99.3 & 73.7 & - & 70.9 & -  \\
Face X-Ray~\cite{li2020xray} & HRNet & Frame & 97.8 & 79.5 & - & 65.5 & 95.4  \\
SLADD~\cite{chen2022SLADD} & Xception & Frame & 98.4 & 79.7 & 76.0 & - & -  \\
SBI*~\cite{shiohara2022sbi} & ResNet50 & Frame & - & 85.7 & - & - & 94.0  \\
UIA-ViT~\cite{zhuang2022uiavit} & ViT-B/16 & Frame & 99.3 & 82.4 & 75.8 & - & 94.7  \\
SeeABLE~\cite{larue2023seeable} & EfficientNet & Frame & - & 87.3 & 86.3 & 75.9 & -  \\
LSDA~\cite{Yan2024LSDA} & EfficientNet & Frame & - & 91.1 & 81.2 & 77.0 & 95.6 \\
\midrule
CNN-GRU~\cite{sabir2019cnngru} & 2D+GRU & Video & 99.3 & 69.8 & - & 68.9 & -  \\ 
FTCN~\cite{zheng2021ftcn} & 3D+TF & Video & 99.8 & 86.9 & - & 74 & -  \\ 
LipForensics~\cite{haliassos2021lips} & 2D+MS-TCN & Video & 99.7 & 82.4 & - & 73.5 & -  \\ 
RealForensics~\cite{haliassos2022realfor} & 2D+CSN & Video & 99.8 & 86.9 & - & 75.9 & -  \\ 
AltFreezing*~\cite{wang2023altfreezing} & 3D & Video & 99.7 & 89.0 & - & - & 93.7  \\ 
TALL~\cite{xu2023tall} & Swin-B & Video & \textbf{99.9} & 90.8 & - & 76.8 & -  \\ 
StyleFlow~\cite{choi2024styleflow} & 3D+GRU+TF & Video & 99.1 & 89.0 & - & - & \textbf{96.1} \\
NACO~\cite{zhang2024naco} & 2D+TF & Video & 99.8 & 89.5 & - & 76.7 & - \\ 
\textcolor{black}{VB-StA~\cite{yan2025generalizing}} & \textcolor{black}{CLIP ViT-B/16} & \textcolor{black}{Video} & - & \textcolor{black}{86.6} & - & \textcolor{black}{77.8} & - \\ 
\textbf{DeepShield} (\textbf{Ours}) & CLIP ViT-B/16+ST-Adapter & Video & 99.2 & \textbf{92.2} & \textbf{93.2} & \textbf{82.8} & \textbf{96.1} \\
\bottomrule
\end{tabular}
\caption{
Comparison results  (video-level AUC, \%)  between DeepShield and existing SOTA algorithms on \textbf{cross-dataset evaluation}.
The comparison methods include both frame-based and video-based approaches, employing various model architectures.
* denotes the results for ~\cite{shiohara2022sbi} and ~\cite{wang2023altfreezing} are taken from~\cite{choi2024styleflow}.
}
\label{tab:maintab}
\end{table*}

\begin{table}[t]
\centering
\scriptsize
\scalebox{0.9}{
\begin{tabular}{c|l|cccc|c}
\toprule
Training Set  & Method &   DF & F2F & FS & NT & Avg. \\ \midrule
 & DCL~\cite{sun2022dcl} & \cellcolor{lightgray!60}\textbf{99.98} & 77.13 & 61.01 & 75.01 & 78.28\\
DF  & UIA-ViT~\cite{zhuang2022uiavit} & \cellcolor{lightgray!60}99.29 & 74.44 & 53.89 & 70.92 & 74.64\\
  & WATCHER~\cite{wang2024watcher} & \cellcolor{lightgray!60}99.68 & 75.50 & 79.51 & 83.66 & 84.59\\
  & \textbf{DeepShield} (\textbf{Ours}) & \cellcolor{lightgray!60}99.94 & \textbf{94.53} & \textbf{80.49} & \textbf{87.28} & \textbf{90.56}\\  \midrule
 & DCL~\cite{sun2022dcl} & 91.91 & \cellcolor{lightgray!60}99.21 & 59.58 & 66.67 & 79.34\\
F2F  & UIA-ViT~\cite{zhuang2022uiavit} & 83.39 & \cellcolor{lightgray!60}98.32 & 68.37 & 67.17 & 79.31\\
  & WATCHER~\cite{wang2024watcher} & 85.73 & \cellcolor{lightgray!60}\textbf{99.82} & 69.38 & 74.16 & 82.27\\
  & \textbf{DeepShield} (\textbf{Ours}) & \textbf{99.76} & \cellcolor{lightgray!60}99.50 & \textbf{83.16} & \textbf{88.63} & \textbf{92.76}\\  \midrule
 & DCL~\cite{sun2022dcl} & 74.80 & 69.75 & \cellcolor{lightgray!60}99.90 & 52.60 & 74.26\\
FS  & UIA-ViT~\cite{zhuang2022uiavit} & 81.02 & 66.30 & \cellcolor{lightgray!60}99.04 & 49.26 & 73.91 \\
  & WATCHER~\cite{wang2024watcher} & 84.52 & 75.06 & \cellcolor{lightgray!60}\textbf{99.95} & 58.81 & 79.59 \\
  & \textbf{DeepShield} (\textbf{Ours}) & \textbf{99.77} & \textbf{95.78} & \cellcolor{lightgray!60}99.58 & \textbf{87.03} & \textbf{95.54}\\  \midrule
 & DCL~\cite{sun2022dcl} & 91.23 & 52.13 & \textbf{79.31} & \cellcolor{lightgray!60}98.97 & 80.41\\
NT  & UIA-ViT~\cite{zhuang2022uiavit} & 79.37 & 67.89 & 45.94 & \cellcolor{lightgray!60}94.59 & 71.95\\
  &WATCHER~\cite{wang2024watcher} & 89.67 & 72.85 & 69.96 & \cellcolor{lightgray!60}98.58 & 82.77\\
  & \textbf{DeepShield} (\textbf{Ours}) & \textbf{99.93} & \textbf{97.33} & 63.31 & \cellcolor{lightgray!60}\textbf{99.03} & \textbf{89.90}\\
  \bottomrule
\end{tabular}
}
\caption{Comparison results  (video-level AUC, \%)  between DeepShield and existing SOTA algorithms on \textbf{cross-manipulation evaluation}.
The results on the diagonal show performance in intra-domain evaluation.
}
\label{tab:cross_mani}
\end{table}

\subsection{Experimental Setups}

\paragraph{Datasets}
To evaluate the generalization capability of the proposed framework, we conduct experiments on the following datasets: FaceForensics++ (FF++)~\cite{rossler2019ff++}, CelebDF v2 (CDF)~\cite{li2020cdf}, DeepFake Detection Challenge (DFDC)~\cite{dfdc}, DFDC Preview (DFDCP)~\cite{dolhansky2019dfdcp}, and Deepfake Detection (DFD)~\cite{dfd}. FF++ consists of 1K real videos and 4K fake videos generated using four manipulation strategies including Deepfake (DF)~\cite{df}, Face2Face (F2F)~\cite{thies2016face2face}, FaceSwap (FS)~\cite{fs} and NeuralTexture (NT)~\cite{thies2019neuraltexture}. 
In our experiments, we use the training split of FF++ (HQ, high-quality version) as the default training set, following recent studies~\cite{haliassos2021lips,wang2023altfreezing,Yan2024LSDA}.

\vspace{-10pt}
\paragraph{Implementation}
In the preprocessing phase, we first utilize RetinaFace~\cite{deng2020retinaface} to detect and crop faces in each frame, and Dlib~\cite{king2009dlib} to detect facial landmarks for generating fake clips. All face crops are then  resized to $224 \times 224$. For both training and testing, we sample four clips of 12 consecutive frames from each video. 
Next, we adopt CLIP ViT-B/16~\cite{CLIP} as the backbone of our framework, with the bottleneck width of ST-Adapter set to 384.
For classification, we implement a binary classifier $\phi$ using a linear layer with softmax activation.
To train the model, we employ the Adam optimizer with a cosine learning rate scheduler over 80 epochs. Specifically, we set the weight decay, initial learning rate, and batch size to $5 \times 10^{-4}$, $3 \times 10^{-4}$, and 16, respectively. Empirically, we set $\omega$ and $\upsilon$ to 0.5 and 0.5 in ~\cref{eq:totalloss}, while 
$\theta$ is set to 10 in ~\cref{eq:patch}. 

\paragraph{Evaluation Metrics}
 Similar to~\cite{haliassos2021lips,zheng2021ftcn,wang2023altfreezing,Yan2024LSDA}, we use Area Under the Curve (AUC) as the evaluation metric and report video-level AUC for fair comparison. For frame-level methods, video-level predictions are obtained by averaging frame-level predictions.

\subsection{Comparisons with the State of the Art}

To assess the generalization of our framework for deepfake detection, we conduct two types of cross-domain evaluations in this study: cross-dataset evaluation and cross-manipulation evaluation. In cross-dataset evaluation, we train on the full FF++ (HQ) dataset and test on four unseen datasets (CDF, DFDCP, DFDC, and DFD), with in-dataset FF++ testing as a baseline. For cross-manipulation evaluation, we train on one of the four FF++ (HQ) manipulation types (DF, F2F, FS, NT) and test on all four for performance comparison.

\vspace{-10pt}

\paragraph{Cross-Dataset Evaluation} 
Table \ref{tab:maintab} illustrates the comparison results between the proposed method and current state-of-the-art (SOTA) algorithms of deepfake video detection in cross-dataset evaluation. 
As showcased, compared to the SOTA algorithms, while our method does not exhibit a significant advantage on the FF++ test set, it achieves the best performance across four unseen datasets. Notably, our approach excels on the most challenging datasets DFDCP and DFDC, outperforming the previous best methods, SeeABLE~\cite{larue2023seeable} and LSDA~\cite{Yan2024LSDA}, by 6.9\% and 5.8\% in AUC, respectively. These findings demonstrate the generalization capacity of the proposed algorithm.

\vspace{-10pt}

\paragraph{Cross-Manipulation Evaluation} 

Table~\ref{tab:cross_mani} reports the comparison results of the cross-manipulation evaluation between DeepShield and three other classic methods for detecting deepfake videos: DCL~\cite{sun2022dcl}, UIA-ViT~\cite{zhuang2022uiavit} and WATCHER~\cite{wang2024watcher}.
The results indicate that our model achieves the highest average performance on both seen and unseen manipulation techniques compared to other detection approaches, consistently excelling across nearly all cases. 
Notably, in the case where FS is used as the training set, our model outperforms the second-best method WATCHER by 15.95\% in AUC on average. In other words, these results underscore the advantage of our method in cross-manipulation evaluation, confirming its strong generalization and effectiveness in deepfake video detection.

\begin{figure}[t]
\centering
\includegraphics[width=1.0\columnwidth]{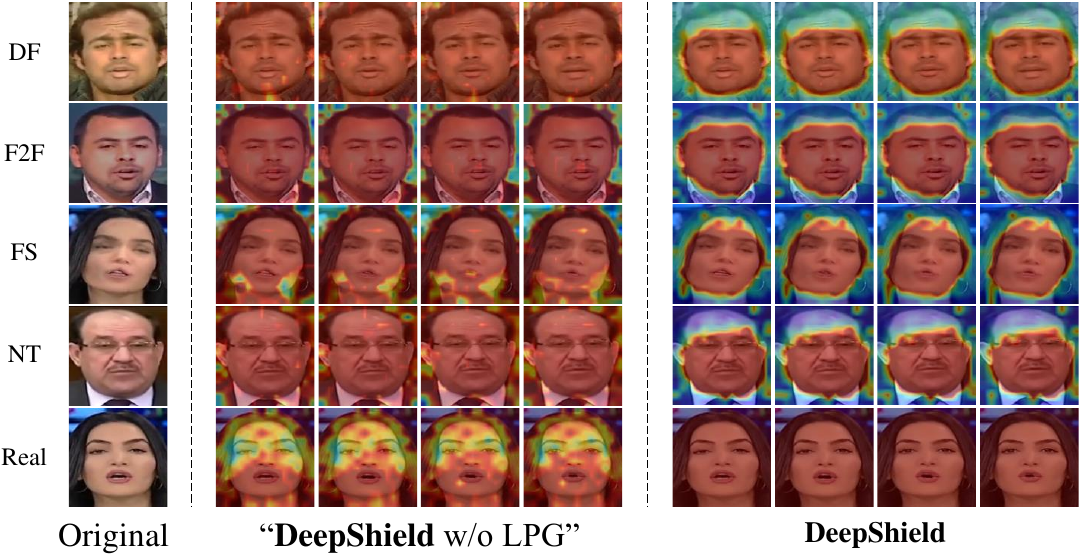} 
\caption{{GradCAM}~\cite{selvaraju2017gradcam} {visualization}  of our proposed DeepShield and its variant ``DeepShield w{/}o LPG''.
We apply Grad-CAM to identify the regions activated for detecting forgery artifacts in both real videos and deepfake videos created with various manipulation techniques. Visualization results are based on intra-dataset scenarios within FF++ (HQ), where warmer colors indicate higher model attention to specific areas.
} 
\label{fig:gradcam}
\end{figure}

\begin{table}[t]
\centering
\scriptsize
\begin{tabular}{l|ccc|c}
\toprule
{Variant}  & CDF  & DFDC & DFDCP & Avg.  \\ 
\midrule
\textbf{DeepShield} (\textbf{Ours})   & \textbf{92.2} & \textbf{82.8} & \textbf{93.2}  & \textbf{89.4} \\
\textbf{DeepShield} w/o LPG      & 89.1 & 81.3 & 87.1  & 85.8 \\
\textbf{DeepShield} w/o GFD $\dagger$       & 89.0 & 81.9 & 91.9  & 87.6  \\
\textbf{DeepShield} w/o LPG \& GFD $\dagger$ $\ddagger$  & 85.4 & 78.4 & 88.9 & 84.2   \\ 
\bottomrule
\end{tabular}
\caption{
Ablation study results (video-level AUC, \%) of DeepShield on both LPG and GFD, showing the impact of removing one or both components.
Here, $\dagger$ indicates that GFD is removed but the model still uses the original real and deepfake video data given in the training set for fine-tuning CLIP-ViT via standard cross-entropy loss. Additionally, $\ddagger$ denotes removal of both LPG and GFD.
These experiments conduct model training on FF++ (HQ) and perform cross-dataset evaluations on CFD, DFDC, and DFDCP. 
}
\label{ablation}
\end{table}

\subsection{Analysis}

\begin{figure}[t]
\centering
\includegraphics[width=0.9\columnwidth]{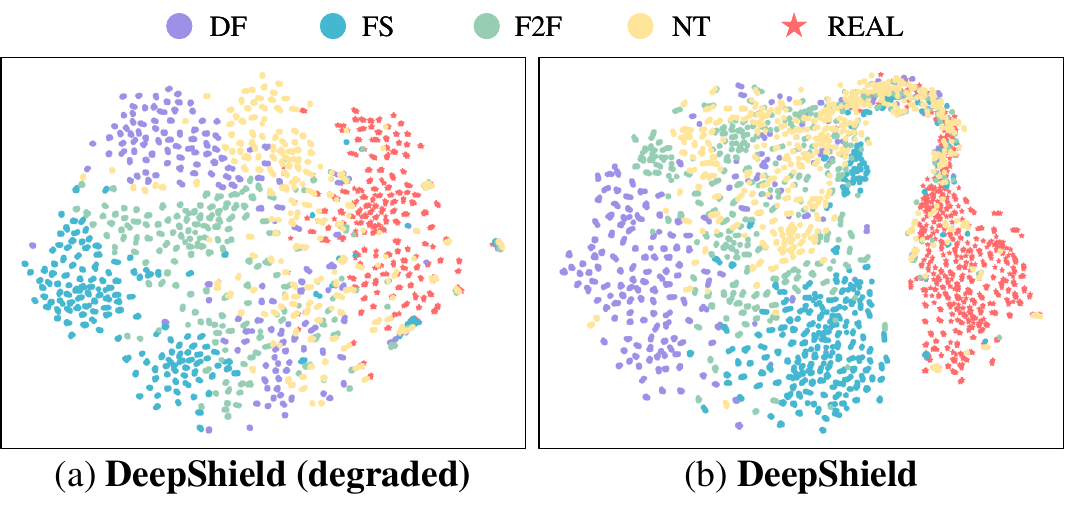} 
\caption{{Feature Visualization} of our proposed DeepShield and its degraded variant using 
{t-SNE}~\cite{van2008tsne}. In the degraded version ``DeepShield (degraded)'', the original CLIP-ViT model is fine-tuned with the ST-Adapter~\cite{pan2022stada} using cross-entropy loss for deepfake video detection.
In this experiment, 8 video clips are sampled from each video in the FF++ test set and encoded their features.
}
\label{fig:tsne} 
\end{figure}

We conduct an ablation study to evaluate the contribution of each component in DeepShield, with results presented in Table~\ref{ablation}, Table~\ref{tab:DFA}, Figure~\ref{fig:gradcam}, and Figure~\ref{fig:tsne}.

\noindent \textbf{How do LPG and GFD Contribute to DeepShield?} 
An ablation study is conducted to evaluate the contributions of LPG and GFD to DeepShield, as summarized in Table~\ref{ablation}, Figure~\ref{fig:gradcam}, and Figure~\ref{fig:tsne}. 
Table~\ref{ablation} first illustrates that 
the removal of LPG leads to a decrease in average AUC from 89.4\% to 85.8\%, highlighting its critical role in improving sensitivity to local forgery traces. Specifically, LPG enables the model to focus on manipulated regions in fake frames and distribute attention more evenly in real frames, as visualized in Figure~\ref{fig:gradcam}. This confirms that LPG enhances the model’s capability to detect subtle manipulation traces. 
Additionally, as shown in Table~\ref{ablation},  the exclusion of GFD causes an AUC reduction of 1.8\%, demonstrating its significance in mitigating forgery-specific overfitting. By synthesizing diverse forgery representations, GFD enhances the model’s adaptability to cross-domain scenarios with varying manipulation styles.
Moreover, it also can be observed in Table~\ref{ablation} that removing both LPG and GFD results in a substantial drop of 5.2\% in average AUC.
As shown in Figure~\ref{fig:tsne}, the full model, with both LPG and GFD, achieves superior feature separation between real and manipulated data, underscoring their combined effectiveness in enhancing generalization and detection performance.

\noindent \textbf{How is the Impact of DFA?}
Here, we investigate how DFA affects the overall detection performance of the proposed model, i.e. DeepShield. To this end, we conduct an ablation study on DFA by removing either DFG or BFG. The experimental results are presented in Table~\ref{tab:DFA}. The table shows that removing either DFG or BFG leads to a decrease in testing detection performance across individual datasets and overall. This highlights the importance of synthesizing diverse and domain-invariant forgery representations for generalized deepfake video detection. Notably, the performance of the two feature synthesis strategies in DFA varies across different datasets when applied individually; only their combined effect achieves optimal results in all datasets.

\begin{table}[t]
\centering
\scriptsize
\begin{tabular}{cc|ccc|c}
\toprule
DFG & BFG & CDF & DFDC & DFDCP & Avg. \\ \midrule
\usym{1F5F8} & \usym{1F5F8} & {\textbf{92.2}} & {\textbf{82.8}} & {\textbf{93.2}} & {\textbf{89.4}}\\
\usym{1F5F8} & - & 91.1 & 81.9 & 92.9 & 88.6\\
- & \usym{1F5F8} & 90.3 & 82.1 & 92.5 & 88.3\\
- & - & 92.0 & 81.3 & 91.6 & 88.3\\
\bottomrule
\end{tabular}
\caption{
Ablation study results (video-level AUC, \%) of DeepShield on DFA, showing the impact of removing its components, DFG and BFG. These experiments conduct model training on FF++ (HQ) and perform cross-dataset evaluations on CFD, DFDC, and DFDCP. 
}
\label{tab:DFA}
\end{table}
\section{Conclusion}
\label{sec:con}

In this work, we propose DeepShield, a novel framework for deepfake video detection that effectively combines local and global analysis to enhance generalization across diverse manipulation techniques. The framework is built upon two complementary components: Local Patch Guidance (LPG) and Global Forgery Diversification (GFD). LPG improves the model's sensitivity to subtle forgery inconsistencies by introducing patch-level supervision, allowing it to detect fine-grained anomalies often overlooked by global methods. Meanwhile, GFD addresses forgery-specific overfitting by synthesizing diverse forgery representations and introducing a novel training strategy with supervised contrastive loss, significantly boosting cross-domain adaptability.
Extensive experiments validate that the synergy of LPG and GFD enables DeepShield to outperform existing methods, particularly in challenging cross-domain settings, demonstrating its robustness for real-world deepfake detection.

\section*{Acknowledgments}
This work is supported in part by the National Key R\&D Program of China (2024YFB3908503), and in part by the
National Natural Science Foundation of China (62322608).

{
    \small
    \bibliographystyle{ieeenat_fullname}
    \bibliography{main}
}

\end{document}


\maketitle

This supplementary material provides additional details and insights into our proposed model namely \textit{DeepShield}. Specifically, we include the following components: detailed model architecture, additional implementation details, and additional experimental analysis. The pseudo-code for the training procedures of DeepShield is illustrated in Algorithm~\ref{alg:train}. 

\begin{algorithm}[ht]
    \caption{Pseudo-code for Training Process}
    \label{alg:train}
    \renewcommand{\algorithmicrequire}{\textbf{Input:}}
    \renewcommand{\algorithmicensure}{\textbf{Output:}}
    
    \begin{algorithmic}[1]
        \REQUIRE 
        Training set of real and fake videos, Training epochs ${\textit{Total\_Epoch}}$, Iterations per epoch ${\textit{Total\_Iter}}$, Initialized model $\psi$ with backbone $E$ and classifier $\phi$
        \ENSURE Optimized model parameters $\psi$
        
        \FOR{$epoch$ = 1 \textbf{to} ${\textit{Total\_Epoch}}$}
            \FOR{$iter$ = 1 \textbf{to} ${\textit{Total\_Iter}}$}
                \STATE Randomly select a mini-batch $\mathcal{B}$ from the training set
                \STATE Initialize $\mathcal{B}^{\text{real}} \leftarrow \emptyset$, $\mathcal{B}^{\text{fake}} \leftarrow \emptyset$, $\mathcal{B}^{\text{blend}} \leftarrow \emptyset$
    
                \FOR{each video $\textbf{V}$ in mini-batch $\mathcal{B}$}
                    \IF{$\textbf{V}$ is real}
                        \STATE Apply SAM to blend video: $\textbf{V}^{\text{blend}} \leftarrow$ SAM($\textbf{V}$)
                        \STATE $\mathcal{B}^{\text{real}} \leftarrow \mathcal{B}^{\text{real}} \cup \{\textbf{V}\}$
                        \STATE $\mathcal{B}^{\text{blend}} \leftarrow \mathcal{B}^{\text{blend}} \cup \{\textbf{V}^{\text{blend}}\}$
                        \STATE $\mathcal{B}^{\text{fake}} \leftarrow \mathcal{B}^{\text{fake}} \cup \{\textbf{V}^{\text{blend}}\}$
                    \ELSE
                        \STATE Randomly select a real video $\textbf{V}^{\text{real}}$ from the training set
                        \STATE $\mathcal{B}^{\text{fake}} \leftarrow \mathcal{B}^{\text{fake}} \cup \{\textbf{V}\}$
                        \STATE $\mathcal{B}^{\text{real}} \leftarrow \mathcal{B}^{\text{real}} \cup \{\textbf{V}^{\text{real}}\}$
                    \ENDIF
                \ENDFOR
                
                \STATE Apply DFA on fake videos: $\mathcal{B}^{\text{DFA}} \leftarrow$ DFA($\mathcal{B}^{\text{fake}}$)
                \STATE $\mathcal{B}^{\text{fake}} \leftarrow \mathcal{B}^{\text{fake}} \cup \mathcal{B}^{\text{DFA}}$
                \STATE $\mathcal{B} \leftarrow \mathcal{B} \cup \mathcal{B}^{\text{DFA}}$
                \STATE $\mathcal{L}^{\text{overall}} \leftarrow \mathcal{L}_{\text{LPG}}(\mathcal{B}^{\text{real}} \cup \mathcal{B}^{\text{blend}}) +  \mathcal{L}_{\text{GFD}}(\mathcal{B})$
                \STATE Optimize $\psi$ using $\mathcal{L}^{\text{overall}}$
            \ENDFOR
        \ENDFOR
        
        \RETURN Optimized model $\psi$
    \end{algorithmic}
\end{algorithm}

\section{Detailed Model Architecture}
\label{sec:model_arch}

\begin{figure}[ht]
\centering
\includegraphics[width=0.75\columnwidth]{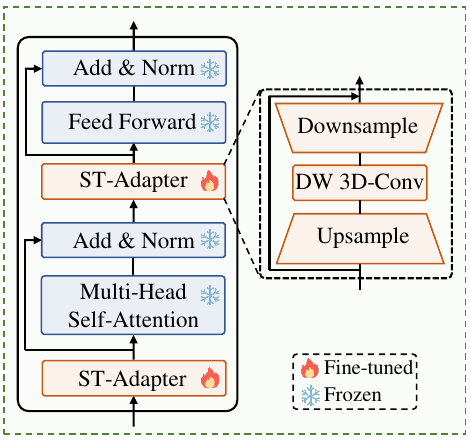} 
\caption{
Detailed architecture of the transformer block in the CLIP image encoder, with two ST-Adapters integrated. Each ST-Adapter incorporates a depth-wise 3D convolution layer to effectively capture temporal information.
} 
\label{fig:model_arch} 
\end{figure}

In this work, we utilize CLIP's pre-trained image encoder, referred to as CLIP-ViT, as the video encoder for generalizable deepfake video detection. To capture spatial manipulations and temporal inconsistencies efficiently, we fine-tune CLIP-ViT with the parameter-efficient ST-Adapter~\cite{pan2022stada}, as depicted in Figure~\ref{fig:model_arch}. The ST-Adapter is inserted before the Multi-Head Self-Attention and Feed-Forward Network in each Transformer block of CLIP-ViT.

Let $\mathbf{X} \in \mathbb{R}^{T \times P \times d}$ denote the input patch embeddings, where $T$ is the number of frames per video clip, $P$ is the number of patch tokens, and $d$ is the embedding dimension.
The ST-Adapter can be defined as follows,
\begin{equation}
  \textrm{ST-Adapter}(\mathbf{X}) = \mathbf{X} + \textrm{g}(\textrm{DWConv3D}(\mathbf{X}\mathbf{W}^{\text{down}}))\mathbf{W}^{\text{up}},
  \label{eq:stada}
\end{equation}
where $\mathbf{W}^{\text{down}}$ and $\mathbf{W}^{\text{up}}$ are learnable parameter weights of the down- and up-projection linear layers, respectively, and $\textrm{g}(\cdot)$ is an activation function. 
As well, $\textrm{DWConv3D}(\cdot)$ is a depth-wise 3D convolution layer with kernel size $T \times H \times W = 3 \times 1 \times 1$, which effectively captures temporal information.

\begin{table*}[t]
\centering
\scriptsize
\begin{tabular}{lccccccc}
\toprule
\multirow{2}{*}{Method} & \multirow{2}{*}{Architecture} & \multirow{2}{*}{Input Type} & \multicolumn{5}{c}{Testing Set AUC (\%)} \\
\cmidrule{4-8}
& & & FF++ & CDF & DFDCP & DFDC & DFD \\
\midrule
CLIP-ViT (vanilla) & ViT-B/16 & Frame & 55.2 & 60.0 & 59.0 & 55.2 & 57.4 \\
CLIP-ViT (adapted) & ViT-B/16+ST-Adapter & Video & 98.2 & 85.4 & 88.9 & 78.4 & 95.3 \\
\textbf{DeepShield} (\textbf{Ours}) & ViT-B/16+ST-Adapter & Video & \textbf{99.2} & \textbf{92.2} & \textbf{93.2} & \textbf{82.8} & \textbf{96.1} \\
\bottomrule
\end{tabular}
\caption{
Comparison results  (video-level AUC, \%)  between DeepShield and its two variants on \textbf{cross-dataset evaluation}.
The comparison methods include both frame-based and video-based approaches, employing various model architectures.
}
\label{tab:varaint}
\end{table*}

\section{Additional Implementation Details}
\label{sec:imple}

Each training batch consists of 16 samples: four real videos, four fake videos (either sampled from the training set or generated using \textit{Spatiotemporal Artifact Modeling}), and eight additional video features derived from these fake videos through \textit{Domain Feature Augmentation} (including \text{Domain-Bridging Feature Generation} and \text{Boundary-Expanding Feature Generation}). During training, four clips of 12 consecutive frames are randomly sampled from each video.
For inference, each video is divided into four segments. From each segment, the first 12 frames are extracted to form four clips, and the final prediction is computed by averaging the prediction probabilities of these clips. 

\begin{table}[t]
\centering
\scalebox{0.7}{
\begin{tabular}{l|cc|c}
\toprule
Method & FaceDancer (WACV23)  & MCNet (ICCV23) & Avg.  \\ 
\midrule
TALL~\cite{xu2023tall}   &  86.0 & 53.3  & 69.7 \\
\textbf{DeepShield} (\textbf{Ours})   & \textbf{98.1} & \textbf{95.3} & \textbf{96.7} \\
\bottomrule
\end{tabular}
}
\caption{
\textcolor{black}{Comparison results of our proposed DeepShield and TALL~\cite{xu2023tall} on the DF40 subset~\cite{yan2024df40} using video-level AUC (\%) as metric.}
}
\label{tab:df40}
\end{table}

\section{Additional Experimental Analysis}

\paragraph{Exploration towards Model Variants}
In this work,  we build two variants of our proposed DeepShield to evaluate its effect on deepfake video detection: \textbf{CLIP-ViT (vanilla)}, which directly employs the pre-trained CLIP image encoder without fine-tuning for the deepfake video detection task, and \textbf{CLIP-ViT (adapted)}, where the original CLIP-ViT model is equipped with the ST-Adapter and fine-tuned via binary real-fake supervision for parameter-efficient adaptation to deepfake video detection. As illustrated by~Table~\ref{tab:varaint} and Table~\textcolor{blue}{1} (of manuscript), our baselines do not demonstrate a significant advantage over existing state-of-the-art (SOTA) methods, ensuring a fair comparison between our full model and others. In the meanwhile, our full model DeepShield consistently outperforms the baseline variants across all cases in cross-dataset evaluations and even surpasses all current SOTA algorithms. This demonstrates the effectiveness of our proposed strategy in significantly improving the model's overall performance in detecting forged samples.

\vspace{-10pt}
\paragraph{More Results on Up-to-date Deepdake Evaluation Dataset}
To better evaluate DeepShield's generalization capability across different types of forgeries, we conducted tests on a subset of the DF40 dataset~\cite{yan2024df40}. This subset includes FaceDancer~\cite{rosberg2023facedancer} (with face-swapping deepfakes generated by neural networks) and MCNet~\cite{hong2023mcnet} (talking-head generation). As shown in Table~\ref{tab:df40}, DeepShield outperforms TALL~\cite{xu2023tall} on these more complex forgeries, demonstrating its strong adaptability even beyond blending-based manipulations.

\vspace{-3pt}

\paragraph{Necessity of Temporal Artifact Generating (TAG)}
In our study, we propose Temporal Artifact Generating (TAG) in the SAM component, which applies Spatial Artifact Generating (SAG) frame-by-frame to simulate temporal inconsistencies in video frames. This strategy maintains the augmentation consistency on source or target frames and uniformity in blending mask adjustments across $T$ frames. 
As shown in Table~\ref{tab:sam_t}, 
removing TAG reduces the average deepfake detection performance by 2.5\%,
 which suggests that using spatial artifact generation alone is insufficient for optimizing DeepShield’s deepfake video detection capabilities. The strong detection performance on both cross-dataset and cross-manipulation cases also confirms that our SAM approach does not cause overfitting, thus maintaining detection efficacy in deepfake video analysis.

\begin{table}[t]
\centering
\scriptsize
\begin{tabular}{l|cccc|c}
\toprule
Variant & CDF  & DFDCP & DFDC &DFD & Avg.  \\ 
\midrule
\textbf{DeepShield} (\textbf{Ours})   & \textbf{92.2} & \textbf{93.2} & \textbf{82.8}  & \textbf{96.1} & \textbf{91.1} \\
\textbf{DeepShield} w/o TAG   &  89.7 & 90.0 & 78.5 & 96.0  & 88.6 \\
\bottomrule
\end{tabular}
\caption{
Ablation study results (video-level AUC, \%) of DeepShield on TAG, showing the impact of removing itself. These experiments conduct model training on FF++ (HQ) and perform cross-dataset evaluations on CFD, DFDC, DFDCP, and DFD. 
}
\label{tab:sam_t}
\end{table}

\paragraph{Hyper-Parameter Sensitivity to the Threshold $\theta$}
We conduct ablation studies to analyze the impact of the threshold $\theta$ in the \textit{Patch Scoring Function} on cross-dataset performance. As shown in Table~\ref{tab:abl_thres}, the threshold plays a critical role in balancing sensitivity and robustness in scoring patch regions. Lower thresholds (e.g., $\theta=10$) result in more consistent performance across datasets, achieving the best average score of 92.7. However, when the threshold becomes too high (e.g., $\theta=150$), the performance slightly declines as critical low-scoring patches may be overlooked. These results demonstrate the importance of carefully tuning $\theta$ to optimize the balance between precision and coverage in patch scoring.
\begin{table}[ht]
\scriptsize
\centering
\begin{tabular}{c|cc|c}
\toprule
{Threshold ($\theta$)} & {CDF} & {DFDCP} & {Avg.} \\ \midrule
10  & 92.2 & \textbf{93.2} & \textbf{92.7} \\
20  & 92.3 & 92.8          & 92.5          \\
50  & \textbf{92.6} & 92.6  & 92.6          \\
100 & 91.9 & 93.0          & 92.4          \\
150 & 92.2 & 92.3          & 92.2          \\ 
\bottomrule
\end{tabular}
\caption{Impact of the threshold $\theta$ in \textit{Patch Scoring Function} on cross-dataset performance, using video-level AUC(\%) as the metric.  
The best results are highlighted in bold.}
\label{tab:abl_thres}
\end{table}

\paragraph{Ablation for Weights in the Losses}
We conduct ablation studies on the balancing weights $\omega$ and $\upsilon$ in the loss function. As shown in Table~\ref{tab:abl_weights}, the weights significantly influence performance. When $\omega = 0.5$ and $\upsilon = 0.5$, the model achieves the best average performance of 89.4, indicating a balanced optimization. Increasing $\omega$ or $\upsilon$ can improve specific dataset scores, such as DFDC (83.1 at $\upsilon = 1$), but may reduce generalization. These results highlight the importance of tuning weights for optimal trade-offs between dataset-specific and overall performance.

\begin{table}[ht]
\centering
\scriptsize
\scalebox{0.8}{
\begin{tabular}{cc|ccc|c}
\toprule
$\omega$ & $\upsilon$   & CDF & DFDC & DFDCP & Avg. \\ \midrule
0.5 & 0.5 & 92.2  & 82.8  & 93.2  & \textbf{89.4}\\
1 & 0.5 & \textbf{92.8}  & 79.4  & \textbf{94.1}  & 88.8\\
5 & 0.5 & 89.8  & 82.0  & 93.1  & 88.3\\
0.5 & 1 & 91.6  & \textbf{83.1}  & 92.9  & 89.2\\
0.5 & 5 & 91.5  & 82.1  & 92.9  & 88.8\\
  \bottomrule
\end{tabular}
}
\caption{Impact of the balancing weights $\omega$ and $\upsilon$ in the loss function on cross-dataset performance, using video-level AUC(\%) as the metric. The best results for each metric are highlighted in bold.}
\label{tab:abl_weights}
\end{table}

{
    \small
    \bibliographystyle{ieeenat_fullname}
    \bibliography{supp}
}
